\documentclass{article}
\usepackage{spconf,amsmath,graphicx}
\usepackage{algorithm}
\usepackage{algorithmic}
\usepackage{comment}
\usepackage{xcolor}
\usepackage{amsmath}
\usepackage{amssymb}
\usepackage{amsfonts}
\usepackage{subfig}
\usepackage{makecell}
\usepackage{arydshln}
\usepackage{booktabs}
\usepackage{caption}
\usepackage[export]{adjustbox}
\usepackage{multirow,makecell,hhline}
\usepackage{soul}

\title{Joint Unsupervised and Supervised Training for Multilingual ASR}
\name{Junwen Bai\sthanks{Work done during an internship at Google.}, Bo Li, Yu Zhang, Ankur Bapna, Nikhil Siddhartha, Khe Chai Sim, Tara N. Sainath}
\address{Google, USA \\
\fontsize{9}{9}\selectfont\ttfamily\upshape
\{junwen,boboli,ngyuzh,ankurbpn,nikhilsid,khechai,tsainath\}@google.com}%
\begin{document}
\ninept
\maketitle
\begin{abstract}
Self-supervised training has shown promising gains in pretraining models and facilitating the downstream finetuning for speech recognition, like multilingual ASR. Most existing methods adopt a 2-stage scheme where the self-supervised loss is optimized in the first pretraining stage, and the standard supervised finetuning resumes in the second stage. In this paper, we propose an end-to-end (E2E) Joint Unsupervised and Supervised Training (JUST) method to combine the supervised RNN-T loss and the self-supervised contrastive and masked language modeling (MLM) losses. We validate its performance on the public dataset \textit{Multilingual LibriSpeech} (MLS), which includes 8 languages and is extremely imbalanced. On MLS, we explore (1) JUST trained from scratch, and (2) JUST finetuned from a pretrained checkpoint. Experiments show that JUST can consistently outperform other existing state-of-the-art methods, and beat the monolingual baseline by a significant margin, demonstrating JUST's capability of handling low-resource languages in multilingual ASR. Our average WER of all languages outperforms average monolingual baseline by 33.3\%, and the state-of-the-art 2-stage XLSR by 32\%. On low-resource languages like Polish, our WER is less than half of the monolingual baseline and even beats the supervised transfer learning method which uses external supervision.
\end{abstract}

\section{Introduction}
\label{sec:intro}

Self-supervised learning is an effective method in unveiling the useful and general latent representations from large-scale unlabeled data. It is often adopted to pretrain a sequence-to-sequence model and facilitate downstream tasks \cite{devlin2018bert,oord2018representation,chen2021injecting}. In speech recognition, recent works have shown successes of the 2-stage pretrain-finetune schemes \cite{wang2020unsupervised,zhang2020pushing,bai2020representation,chung2021w2v,hsu2021hubert}. Pretrained models can greatly reduce the sample complexity for downstream finetuning. For instance, finetuning wav2vec 2.0 (w2v2) pretrained on 60k hours with only 1h labeled data can outperform most fully supervised models \cite{baevski2020wav2vec}.

While self-supervised learning has been successful for sequence modeling, some concerns have also been raised. For example, finetuning a pretrained model is prone to catastrophic forgetting \cite{chen2020recall,kessler2021continual}. The model might \textit{forget} the previously learnt knowledge when trained with supervision, particularly when the supervised set is large. Another concern is the pretrained checkpoint selection. The downstream performance varies from one checkpoint to another, and the one pretrained longer may not be the best one. These issues are even more severe in multilingual ASR, since different languages are often heterogeneous and the corpus is often imbalanced. In \textit{Multilingual LibriSpeech} (MLS) \cite{pratap2020mls}, English has up to 44k hours while Polish only has 100 hours. Most existing methods tackle multilingual ASR from 2 directions. The first direction is transfer learning from a source multilingual corpus to a target low-resource multilingual dataset. In \cite{li2021scaling}, the work first trains the model on Google's 15-language VoiceSearch (VS) traffic and then uses it to seed the transfer learning on MLS. Even though some languages in MLS 
are not included in VS, the model can deliver satisfactory WERs on those low-resource languages, demonstrating its generalization capability. However, such transfer learning requires massive supervised source corpora which may not be easily accessible. Another direction is to learn useful representations through pretraining and perform finetuning with supervision, similar to monolingual ASR. 
\cite{lample2019cross} explores the unsupervised pretraining using cross-lingual language modeling and \cite{riviere2020unsupervised} investigates the cross-lingual transfer of phoneme features.
XLSR \cite{conneau2020unsupervised} builds on w2v2 and pretrains the model on 53 languages using the self-supervised losses. XLSR also stands for the state-of-the-art (SOTA) on MLS dataset. 

In this paper, we propose a novel Joint Unsupervised and Supervised Training (JUST) method for multilingual ASR, to reconcile the unsupervised and supervised losses synergistically. JUST includes two self-supervised losses, contrastive loss \cite{baevski2020wav2vec} and MLM loss \cite{devlin2018bert}, together with a supervised RNN-T loss \cite{graves2012sequence}. Our model architecture inherits from w2v-bert \cite{chung2021w2v}, a novel variation of w2v2. The outputs from w2v-bert are passed to the decoder and produce the RNN-T loss. We explore two types of learning with JUST: 1) JUST trained from scratch, and 2) JUST finetuned from a pretrained checkpoint. We compare these 2 settings with XLSR and other standard baselines. Experiments show that JUST can consistently outperform other SOTA and baselines. For instance, on 8 languages from MLS, JUST improves over XLSR by 30\% on average. 
\section{Related Work}
Early works adopted joint training to learn robust and transferable representations. In NLP, \cite{cheng2021self} proposes joint training for machine translation. \cite{gururangan2020don} suggests multiple pretraining objectives for domain-adaptive applications. In speech, PASE \cite{pascual2019learning} jointly solves multiple self-supervised tasks to learn general representations. More recent research found the joint training with both supervised and unsupervised losses can directly optimize the ASR performance. \cite{talnikar2021joint} alternatively minimizes an unsupervised masked CPC loss and a supervised CTC loss \cite{graves2006connectionist}. This single-stage method is shown to match the performance of the two-stage w2v2 on the Librispeech 100-hours dataset. Similarly, UniSpeech \cite{wang2021unispeech} optimizes a combination of phonetic CTC loss and contrastive loss. To further increase the quantizer codebook usage, UniSpeech randomly replaces contextual representations with quantized latent codes. \cite{raghavanhybrid} also designs a similar hybrid multitask learning to train acoustic models under low-resource settings, comprising of supervised CTC, attention and self-supervised reconstruction losses. Similarly, \cite{hwang2021large} combines self- and semi-supervised learning methods for online ASR model. These methods only contain one self-supervised loss in their optimization and often tackle with speech recognition in the phoneme level \cite{wang2021unispeech,raghavanhybrid}. JUST incorporates two self-supervised losses (contrastive and MLM losses), and replaces the CTC loss with an RNN-T loss. RNN-T extends CTC with a prediction network to simulate the effect of LM and has been widely adopted in prior multilingual ASR systems \cite{li2021scaling}.
Furthermore, unlike \cite{pascual2019learning,raghavanhybrid} where each of the multiple tasks has its own branch, JUST computes different losses simply using the intermediate outputs from different layers (Fig.~\ref{fig:just}).  
\section{Method}

\begin{figure}[t]
	\centering
	\includegraphics[width=0.41\textwidth]{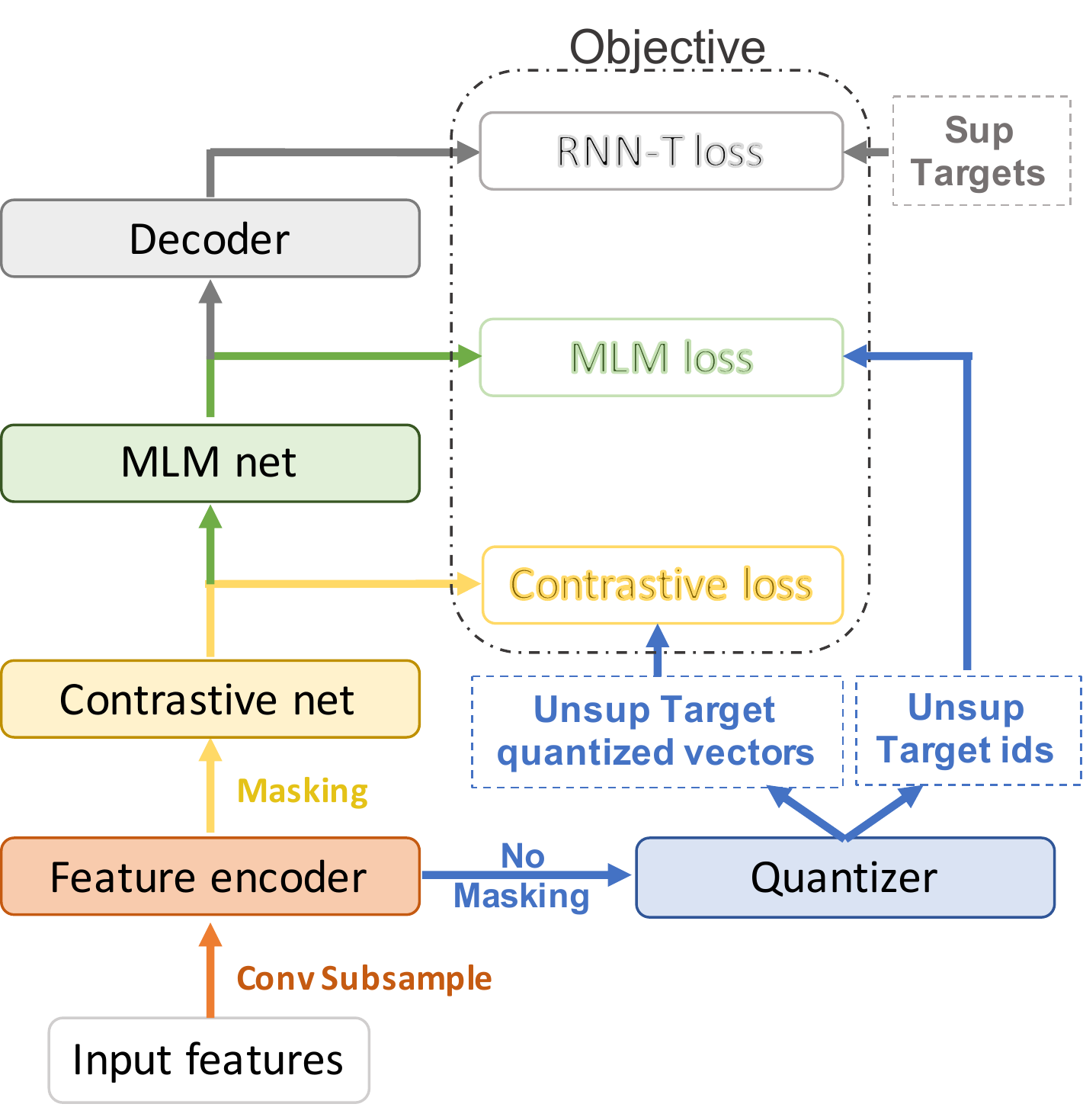}
    \caption{An overview of our JUST framework. Feature encoder, contrastive net, MLM net and decoder are stacked sequentially. The output of each module constitutes a loss in the objective function. Target vectors and ids in the blue boxes are for unsupervised losses. Supervised targets in the grey box are for RNN-T loss.
    }
	\label{fig:just}
\end{figure}

Our JUST framework is comprised of multiple modules for unsupervised and supervised losses. All modules (except for the quantizer) are stacked sequentially and each reads the output from the previous module. We will elaborate on them, along with the losses, in the following sections. Fig.~\ref{fig:just} presents an overview. 

\subsection{Feature encoder}

The feature encoder converts the original log-mel filter bank features $\{x_i\}_{i=1}^L$ to the latent speech representations $\{z_i\}_{i=1}^T$ for $T$ time steps. $T$ is smaller than the original length $L$ due to time reduction.
Unlike \cite{baevski2020wav2vec} where seven blocks of CNN are used, JUST only has two CNN blocks both with filter size 3$\times$3 and strides (2, 2), the same as \cite{zhang2020pushing}. One can also view the feature encoder as a convolutional subsampling, with 4x reduction in the feature dimensionality and sequence length.

\subsection{Quantizer}
JUST adopts a complex quantization mechanism \cite{baevski2020wav2vec}. After the abstraction of the original inputs through feature encoder, the latent representations $\{z_i\}_{i=1}^T$ are passed to a quantizer (without any masking). The goal of the quantizer is to ``summarize" all the latent speech representations to a finite set (referred to as a codebook) of representative discriminative speech tokens $\{e_j\}_{j=1}^V$ where $V$ is the size of the codebook. The codebook in the quantizer stores all these tokens and each latent representation from the feature encoder is mapped to a token index corresponding to a token in the codebook, through Gumbel softmax \cite{jang2016categorical} which enables differentiation of discrete codebook selection. JUST uses a single codebook rather than multiple ones \cite{baevski2020wav2vec}. All the tokens in the codebook are learnable during training. Quantizer module generates target quantized vector (token) $q_i$ and target id (token index) $y_i$ for each $z_i$, where $q_i\in\{e_j\}_{j=1}^V$, $y_i\in[1...V]$. To encourage the use of the codebook, \cite{baevski2020wav2vec} introduces the entropy-based diversity loss $\mathcal L_d$. We include it in JUST as well. 

\subsection{Contrastive net}

The outputs of feature encoder $\{z_i\}_{i=1}^T$ are not only used for quantization, but also fed into the contrastive net after masking. For masking, some $z_i$'s are randomly chosen and replaced with random vectors. Contrastive net reads $z_i$ of all time steps (either masked or unmasked) and outputs contrastive context vectors $\{c_i\}_{i=1}^T$ for deriving contrastive self-supervised loss. Contrastive net is a stack of Conformer blocks \cite{gulati2020conformer}, each with multi-headed self-attention, depth-wise convolution and feed-forward layers. To derive the contrastive loss $\mathcal L_c$, for anchor $c_i$, we take $q_i$ as the positive sample and $K$ negative samples/distractors $\{\tilde{q}_i\}_{i=1}^K$ uniformly sampled from $q_j$ of other masked $z_j$'s in the same utterance: 
\begin{equation}
\mathcal L_c=-\log \frac{sim(c_i, q_i)}{sim(c_i,q_i)+\sum_{j=1}^Ksim(c_i,\tilde{q}_j)}
\end{equation}
where $sim(a,b)$ is the exponential of the cosine similarity between $a$ and $b$.

\subsection{MLM net}

We further boost the contextualized representation learning through a masked prediction task with the quantizer. The inputs of MLM net are $\{c_i\}_{i=1}^T$ from contrastive net. Similar to contrastive net, MLM net is also a stack of Conformer blocks. We denote the outputs of MLM net as $\{m_i\}_{i=1}^T$, which are high-level context vectors. Each $m_i$ is used for token id prediction through a linear layer. The predicted id $\hat{y}_i\in[1..V]$ is compared with the target token id $y_i$ from the quantizer, by the standard cross-entropy loss $\mathcal L_{m}$. 

Together with $\mathcal L_{d}$ and $\mathcal L_{c}$, the unsupervised loss is computed as:
\begin{equation}
\mathcal L_u = \mathcal L_c+\mathcal L_m+\alpha \mathcal L_d
\end{equation}
$\alpha$ is set to 0.1 following \cite{chung2021w2v}.

\begin{table*}[t]
\centering
\begin{tabular}{lc@{\hspace{0.50em}}c@{\hspace{0.50em}}c@{\hspace{0.50em}}c@{\hspace{0.50em}}c@{\hspace{0.50em}}c@{\hspace{0.50em}}c@{\hspace{0.50em}}c@{\hspace{0.50em}}c@{\hspace{0.50em}}c@{\hspace{0.50em}}c@{\hspace{0.50em}}}
\toprule 
Method & External data &\textbf{de} & \textbf{en} & \textbf{es} & \textbf{fr} & \textbf{it} & \textbf{nl} & \textbf{pl} & \textbf{pt} & Avg & Avg (w/o \bf{en}) \\
\midrule
\midrule
Monolingual \cite{pratap2020mls} & - & 7.10 & 6.76 & 6.68 & 6.58 & 11.78 & 13.09 & 21.66 & 20.52 & 11.8 & 12.5\\
\quad + 5-gram LM \cite{pratap2020mls} & - & 6.49 & 5.88 & 6.07 & 5.58 & 10.54 & 12.02 & 20.39 & 19.49 & 10.8 & 11.5\\
\midrule
XLSR-53 \cite{conneau2020unsupervised} & Y & 7.0 & - & 6.3 & 7.6 & 10.4 & 10.8 & 17.2 & 14.7 & 10.6 & 10.6\\
B0 (random init.) \cite{li2021scaling} & Y & 5.5 & 6.1 & 5.8 & 6.9 & 11.9 & 11.9 & 15.4 & 16.2 & 10.0 & 10.5\\
B0 (15-language model init.) \cite{li2021scaling} & Y & 5.0 & 6.6 & 4.7 & 6.1 & 10.1 & 11.1 & 10.9 & 15.5 & 8.8 & 9.1\\
E3 (15-language model init.) \cite{li2021scaling} & Y & 4.3 & \textbf{5.8} & 4.2 & \textbf{4.9} & 8.8 & 9.9 & 10.4 & 15.2 & 7.9 & 8.2 \\
\midrule
JUST ($\beta=0$) & N & 5.5 & 6.9 & 4.1 & 6.0 & 9.3	& 10.3 & 11.3 & 9.4 & 7.8 & 8.0 \\
w2v2 Pretrain ($\mathcal L_c$+$\alpha\mathcal L_d$)+pure Finetune ($\mathcal L_s$) & N & 4.7 & 6.8 & 4.1 & 5.8 & 9.9 & 10.3 & 12.1 & 12.6 & 8.3 & 8.5 \\
w2v-bert Pretrain ($\mathcal L_u$)+pure Finetune ($\mathcal L_s$) & N & 4.3 & 6.6 & 3.8 & 5.0 & 9.1 & 9.9 & 8.1 & 14.6 & 7.7 & 7.8 \\
w2v-bert Pretrain ($\mathcal L_u$)+JUST Finetune ($\mathcal L$) & N & 4.2 & 6.6 & 4.0 & 5.0 & 9.0 & 9.5 & 7.6 & 15.1 & 7.6 & 7.8 \\
w2v2 Joint Training ($\mathcal L_s$+$\beta$($\mathcal L_c$+$\alpha\mathcal L_d$)) & N & 4.6 & 6.7 & 4.1 & 5.7 & 8.9 & 9.9 & 9.8 & 9.3 & 7.4 & 7.5 \\
JUST ($\mathcal L$) & N & 4.6 & 6.8 & 3.9 & 5.7 & 9.1 & 9.9 & 9.1 & 8.6 & 7.2 & 7.3 \\
JUST ($\mathcal L$)+pure Finetune ($\mathcal L_s$) & N & \textbf{4.1} & 6.5 & \textbf{3.7} & 5.2 & \textbf{8.2} & \textbf{9.5} & \textbf{6.6} & \textbf{8.0} & \textbf{6.5} & \textbf{6.5} \\
\bottomrule
\end{tabular}
\caption{WER(\%) results on MLS for different methods. JUST-based methods greatly outperforms the compared baselines. XLSR-53 used external unsupervised data for pretraining. B0 and E3 used external supervised data. Our JUST did not use any external data.}
\label{tbl:mls}
\end{table*}

\begin{table*}[t]
\centering
\begin{tabular}{lcccccccccccc}
\toprule 
Method & Ext. &\textbf{de} & \textbf{en} & \textbf{es} & \textbf{fr} & \textbf{it} & \textbf{nl} & \textbf{pl} & \textbf{pt} & Avg & Avg (w/o \bf{en}) \\
\midrule
\midrule
JUST ($\beta=0$) & N & 5.5 & 6.9 & 4.1 & 6.0 & 9.3	& 10.3 & 11.3 & 9.4 & 7.8 & 8.0 \\
JUST ($\beta=0.03$) & N & 5.0 & 7.4 & 4.3 & 6.3 & 9.3 & 10.3 & 8.7 & 9.1 & 7.5 & 7.6 \\
JUST ($\beta=0.05$) & N & 5.2 & 6.8 & 4.4 & 5.7 & 9.4 & 9.9 & 9.3 & 8.8 & 7.4 & 7.5 \\
JUST ($\beta=0.07$) & N & 4.6 & 6.8 & 3.9 & 5.7 & 9.1 & 9.9 & 9.1 & 8.6 & 7.2 & 7.3 \\
JUST ($\beta=0.1$) & N & 5.8 & 6.8 & 4.1 & 5.8 & 10.3 & 10.0 & 9.7 & 8.6 & 7.6 & 7.8 \\
\bottomrule
\end{tabular}
\caption{Weight sensitivity study on $\beta$. When $\beta=0.07$, the unsupervised loss is roughly the same as the supervised loss, which means balancing the unsupervised and supervised losses can be critical in joint training.}
\label{tbl:beta_sensitivity}
\vspace{-1em}
\end{table*}

\subsection{Decoder}

The decoder of JUST is a 2-layer RNN Transducer. $\{m_i\}_{i=1}^T$ are passed through Swish activation, batch normalization, and finally fed into the decoder. The output vocabulary of the decoder is a unified grapheme set pooled from all the 8 languages in MLS. RNN-T loss is used in this work as the supervised loss, denoted by $\mathcal L_s$. Our final objective function is simply the combination of $\mathcal L_u$ and $\mathcal L_s$:
\begin{equation}
\mathcal L = \mathcal L_s + \beta \mathcal L_u
\end{equation}
$\beta$ is a trade-off weight. $\mathcal L$ is optimized via Adam \cite{kingma2014adam}.

\section{Experiments}

\subsection{Dataset}

MLS dataset \cite{pratap2020mls} is used as the benchmark in our experiments. It is derived from read audiobooks of LibriVox. There are 8 languages (namely English ({\bf en}), German ({\bf de}), Dutch ({\bf nl}), French ({\bf fr}), Spanish ({\bf es}), Italian ({\bf it}), Portuguese ({\bf pt}) and Polish ({\bf pl})), with 44.5k hours of English and 6k hours for other languages combined. Some low-resource language like Polish only has 100 hours. Each utterance is 10-20 seconds long.

\subsection{Training details}
\noindent\textbf{Architecture~} The inputs are 80-d filter bank features. Feature encoder has 2 convolutional layers with filter size (3,3), strides (2,2). Two layers have 128 and 32 channels respectively. Contrastive net consists of 8 Conformer blocks, each with hidden dimensionality 1024, 8 attention heads and convolution kernel size 5. MLM net consists of 16 Conformer blocks with the same configuration. Our decoder uses a 2-layer 768-d LSTM-based RNN-T with 3072 hidden units. 
$\{c_i\}_{i=1}^T$ from contrastive net and $\{m_i\}_{i=1}^T$ from MLM net are used in computing self-supervised losses after layer normalization.
Our codebook has size $V$=1024, with each token of length 1024.

\noindent\textbf{Masking~} To mask $\{z_i\}_{i=1}^T$, we randomly sample 6.5\% of all time steps and replace each of the selected time steps and its subsequent 10 time steps with random normal vectors (from $\mathcal N(0,0.1)$). Some spans might overlap. 

\noindent\textbf{Hyperparameters~} We train JUST with batch size 1024 on 64 TPUs. Adam optimizer is employed with $\beta_1=0.9,\beta_2=0.98$ for training. Our global learning rate schedule is the same as \cite{chung2021w2v} but with warm-up steps 5000 and peak learning rate 4e-4. Decoder uses a separate schedule rather than the global one, with warm-up steps 1500 and peak learning rate 7e-4. We set $\alpha=0.1$ following prior works \cite{zhang2020pushing,chung2021w2v} and $\beta=0.07$ via tuning with grid search. 

\noindent\textbf{Evaluation~} We show the WER for each language, as well as the average WER with or without \textbf{en} included.

\subsection{Compared methods}

We compare JUST with several baselines. MLS paper \cite{pratap2020mls} provides competitive monolingual baselines without any LM and with a 5-gram LM. Using LM improves the monolingual performance. XLSR \cite{conneau2020unsupervised} pretrains a w2v2 on 53 languages from MLS, CommonVoice and BABEL, and finetunes the model on MLS. XLSR finetuned on the full set of MLS can outperform some low-resource monolingual baselines like \textbf{it}, \textbf{pt}, \textbf{pl}, but not all (Table~\ref{tbl:mls}). We also include transfer learning models, B0 and E3, from \cite{li2021scaling}, which used heavy supervision from external \textit{VoiceSearch} (VS) dataset containing 15 languages. Both B0 and E3 are first trained on VS with supervision and then finetuned on MLS. B0 is a smaller model with 370M parameters and E3 is a larger model with 1B parameters. We also include a B0 model trained from scratch for comparison. Besides these existing baselines from literature, we further train a w2v-bert model from scratch on MLS (JUST with $\beta=0$), and a 2-stage preatrain-finetune w2v-bert model on MLS without any external data.

\begin{figure*}[h!]
  \centering
  \begin{minipage}{0.33\textwidth}
    \centering
    \includegraphics[width=0.88\textwidth]{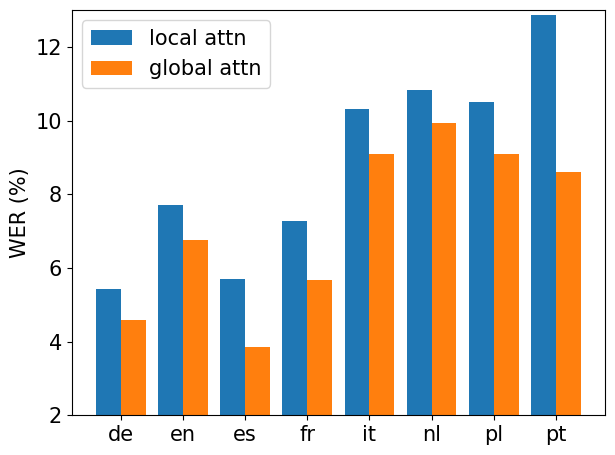}
  \end{minipage}
  \hfill
  \begin{minipage}{0.33\textwidth}
    \centering
    \includegraphics[width=0.88\textwidth]{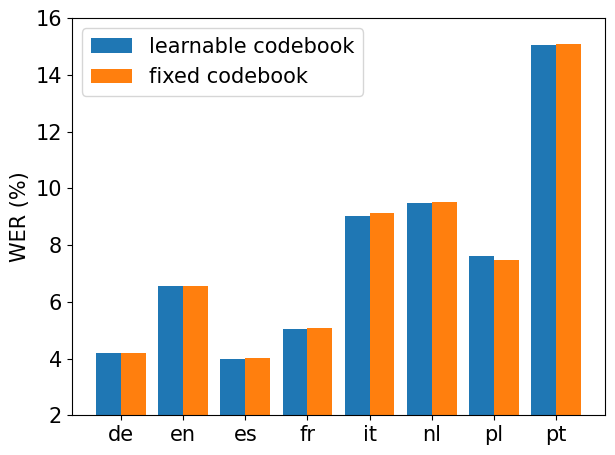}
  \end{minipage}
  \hfill
  \begin{minipage}{0.33\textwidth}
    \centering
    \includegraphics[width=0.88\textwidth]{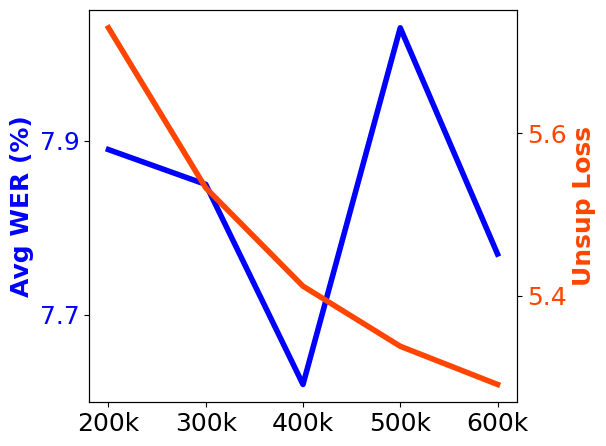}
  \end{minipage}
  \caption{\textbf{Left:} Comparison between using local attention with left/right context size 128, and using global attention. Global attention clearly boosts the performance. \textbf{Middle:} For JUST finetuning, we either allow the codebook to be updated, or to remain fixed during the whole finetuning. They deliver similar results. \textbf{Right:} The average WERs of JUST finetuning ($\beta=0.01$) from different checkpoints. The checkpoint with smaller unsupervised loss may not lead to the best finetuning results.}
  \label{fig:attn_cb}
\end{figure*}

\subsection{Results}

For JUST, we either train it from scratch on MLS, or jointly finetune it from a pretrained checkpoint on MLS where the pretraining phase would only optimize the unsupervised loss. Note that compared to XLSR, our pretraining would incorporate more self-supervised losses. Our JUST has 600M parameters, which is roughly the same scale as B0, XLSR but much smaller than E3. 

\noindent\textbf{Average WER~} On the average WER of all 8 languages, all JUST-based methods outperform previous works. In particular, JUST (with $\beta=0.07$) outperforms the monolingual baseline with 5-gram LM by 33.3\%, XLSR-53 by 32.0\%, B0 by 18.2\%, E3 by 8.8\%. Note that E3 is a transfer learning method with much larger size and heavy external supervision. JUST's improvement over E3 validates the effectiveness of our architecture and joint training scheme. If we exclude English WER and compare other languages as in XLSR \cite{conneau2020unsupervised}, JUST outperforms monolingual, XLSR-53, B0, E3 by 36.5\%, 31.1\%, 19.8\%, 11.0\% respectively. Compared to JUST with $\beta=0$, JUST with joint training improves the average WER (w/o \textbf{en}) by 7.7\% (8.8\%). To show the necessity of MLM loss in joint training, we further include the results of w2v2 joint training with the objective $\mathcal L_s$+$\beta$($\mathcal L_c$+$\alpha\mathcal L_d$). JUST still performs better on the average WERs and low-resource \textbf{es}, \textbf{pl}, \textbf{pt}.

\noindent\textbf{Low-resource languages~} JUST improves WERs for low-resource languages such as \textbf{pl}, \textbf{pt}. On \textbf{pl}, JUST's WER is less than half of the monolingual WER baseline and roughly half of XLSR's WER. On \textbf{pt}, JUST's WER is at least 40\% lower than any of XLSR, B0 or E3, which is significant. 

\noindent\textbf{JUST finetune~} Two finetuning schemes are attempted. \textbf{First}, we take a pretrained checkpoint trained with $\mathcal L_u$, and finetune it with JUST objective $\mathcal L$. Compared to w2v2 Pretrain+pure Finetune (no MLM loss), it improves on all languages except \textbf{pt}. Compared to w2v-bert Pretrain+pure Finetune (with MLM loss), it also improves on \textbf{de}, \textbf{en}, \textbf{fr}, \textbf{it}, \textbf{nl}, \textbf{pl}. It is interesting to compare JUST and JUST Finetune on \textbf{pt}, \textbf{pl}. Different training schemes lead to different quantized tokens and cause the discrepancy. Empirically, JUST from scratch can better facilitate the low-resource languages and reduce WER of each language to below 10. For JUST finetuning, we set $\beta=0.01$ to de-weight the unsupervised loss instead of matching with the supervised loss.
\textbf{Second}, we take a checkpoint from JUST trained from scratch, and finetune it with only supervised loss $\mathcal L_s$. It achieves the best average WER, further improving the average WER (w/o \textbf{en}) of JUST by 10\% (11\%). On \textbf{de}, \textbf{es}, \textbf{it}, \textbf{nl}, \textbf{pl}, \textbf{pt}, this scheme outperforms all compared methods and remains competitive on other languages.

\noindent\textbf{$\beta$ sensitivity~} Table~\ref{tbl:beta_sensitivity} also includes the sensitivity study on $\beta$. When $\beta=0.07$, the unsupervised and the supervised losses are balanced, resulting in the best performance.

\noindent\textbf{Attention~} We compare two attention mechanisms for JUST from scratch: a local attention mechanism with both left and right context 128, and a global attention mechanism with full context. The results are shown in Fig.~\ref{fig:attn_cb}. Global attention clearly outperforms local attention on all languages.

\noindent\textbf{Codebook~} Original w2v2 doesn't update codebook in the finetuning phase. JUST finetuning, however, keeps the unsupervised loss and could further update the codebook. We compare the performance with learnable or fixed codebook during JUST finetuning ($\beta=0.01$), and find their results are close (Fig.~\ref{tbl:beta_sensitivity}). This implies that fixing codebook in JUST finetuning would not degrade the performance. In practice, we also find larger $\beta$ would bias the updates to the codebook, leading to worse results.

\noindent\textbf{Pretrained checkpoints~} Different checkpoints can lead to different downstream performance. The later checkpoints do not necessarily lead to better downstream WERs. To verify this, we finetune multiple pretrained checkpoints and evaluate their finetuning quality. The rightmost subfigure from Fig.~\ref{fig:attn_cb} shows the constantly descending unsupervised loss $\mathcal L_u$, while the downstream average WERs don't follow the same trend.

\section{Conclusion}

This work proposes a novel uniform multilingual ASR system for the end-to-end speech recognition on multiple languages. Our method, JUST, is composed of a contrastive module for learning discrete speech representations and an MLM module that performs a masked language modeling task. 
JUST jointly optimizes the unsupervised contrastive loss and MLM loss, together with the supervised RNN-T loss. Compared to the prevalent 2-stage pretrain-finetune models, JUST-based methods can guide the whole training process with the unsupervised and supervised losses jointly. JUST's performance is validated on a public multilingual ASR dataset, MLS, and outperforms the monolingual baselines, a SOTA 2-stage pretrain-finetune model XLSR, and the latest transfer learning methods, proving  the effectiveness of joint training. On low-resource languages, our JUST and its variants can consistently bring gains and boost performance. In the future, we will investigate how the objective function affects the codebook learning, and also explore the joint training with more languages and other unsupervised losses, as well as the tradeoff between unsupervised and supervised components. 

\section{Acknowledgments}

We would like to thank Yu-An Chung, Yotaro Kubo and Weiran Wang for constructive suggestions.

\clearpage
\bibliographystyle{IEEEbib}
\bibliography{refs}

\end{document}